\documentclass[letterpaper]{article} 
\usepackage{aaai25}  
\usepackage{times}  
\usepackage{helvet}  
\usepackage{courier}  
\usepackage[hyphens]{url}  
\usepackage{graphicx} 
\usepackage{natbib}  
\usepackage{caption} 
\usepackage{algorithm}
\usepackage{algorithmic}
\usepackage{amsmath}
\usepackage{newfloat}
\usepackage{listings}
\usepackage{url}
\usepackage{multirow}
\usepackage{graphicx}
\usepackage{subcaption}
\newtheorem{hyp}{Hypothesis}
\DeclareCaptionStyle{ruled}{labelfont=normalfont,labelsep=colon,strut=off}

\floatstyle{ruled}
\newfloat{listing}{tb}{lst}{}
\floatname{listing}{Listing}

\title{Devil in the Tail: A Multi-Modal Framework for Drug-Drug Interaction Prediction in Long Tail Distinction}

\author{
Liangwei Nathan Zheng\textsuperscript{1}, 
Chang George Dong\textsuperscript{1},
Wei Emma Zhang\textsuperscript{1}, 
Xin Chen\textsuperscript{2}, 
Lin Yue\textsuperscript{1}, 
Weitong Chen\textsuperscript{1} \\
\{liangwei.zheng, chang.dong, wei.e.zhang, lin.yue, t.chen\}@adelaide.edu.au, xin.chen@nottingham.ac.uk
}
\affiliations {
The University of Adelaide\textsuperscript{1}, The University of Nottingham\textsuperscript{2}
}

\usepackage{bibentry}

\begin{document}

\maketitle

\begin{abstract}
    Drug-drug interaction (DDI) identification is a crucial aspect of pharmacology research. There are many DDI types (hundreds), and they are not evenly distributed with equal chance to occur. Some of the rarely occurred DDI types are often high risk and could be life-critical if overlooked, exemplifying the long-tailed distribution problem. Existing models falter against this distribution challenge and overlook the multi-faceted nature of drugs in DDI prediction. In this paper, a novel multi-modal deep learning-based framework, namely TFDM, is introduced to leverage multiple properties of a drug to achieve DDI classification. The proposed framework fuses multimodal features of drugs, including graph-based, molecular structure, Target and Enzyme, for DDI identification. To tackle the challenge posed by the distribution skewness across categories, a novel loss function called Tailed Focal Loss is introduced, aimed at further enhancing the model performance and address gradient vanishing problem of focal loss in extremely long-tailed dataset. Intensive experiments over 4 challenging long-tailed dataset demonstrate that the TFMD outperforms the most recent SOTA methods in long-tailed DDI classification tasks. The source code is released to reproduce our experiment results: \url{https://github.com/IcurasLW/TFMD_Longtailed_DDI.git}
\end{abstract}

\section{Introduction}
With the rapid increase of drug types, thousands of interactions between drugs have been investigated including adverse and coupling interactions. Unfortunately, a multitude of patients have suffered adverse outcomes, including severe harm and even death, due to issues arising from drug dispensation and a lack of awareness regarding drug-drug interactions (DDIs)\cite{tariq2018medication}. Furthermore, the process of investigating the interaction between drugs is time-consuming, financially burdensome, and labour-intensive in nature\cite{yeh2014detection,yue2020deep,li2023artificial}. This limitation in identifying new DDIs also happens with the advancement of medication. Thus, a robust and precise computational method is expected to predict the potential DDIs.

\begin{figure}
    \centering
    \includegraphics[scale=0.18]{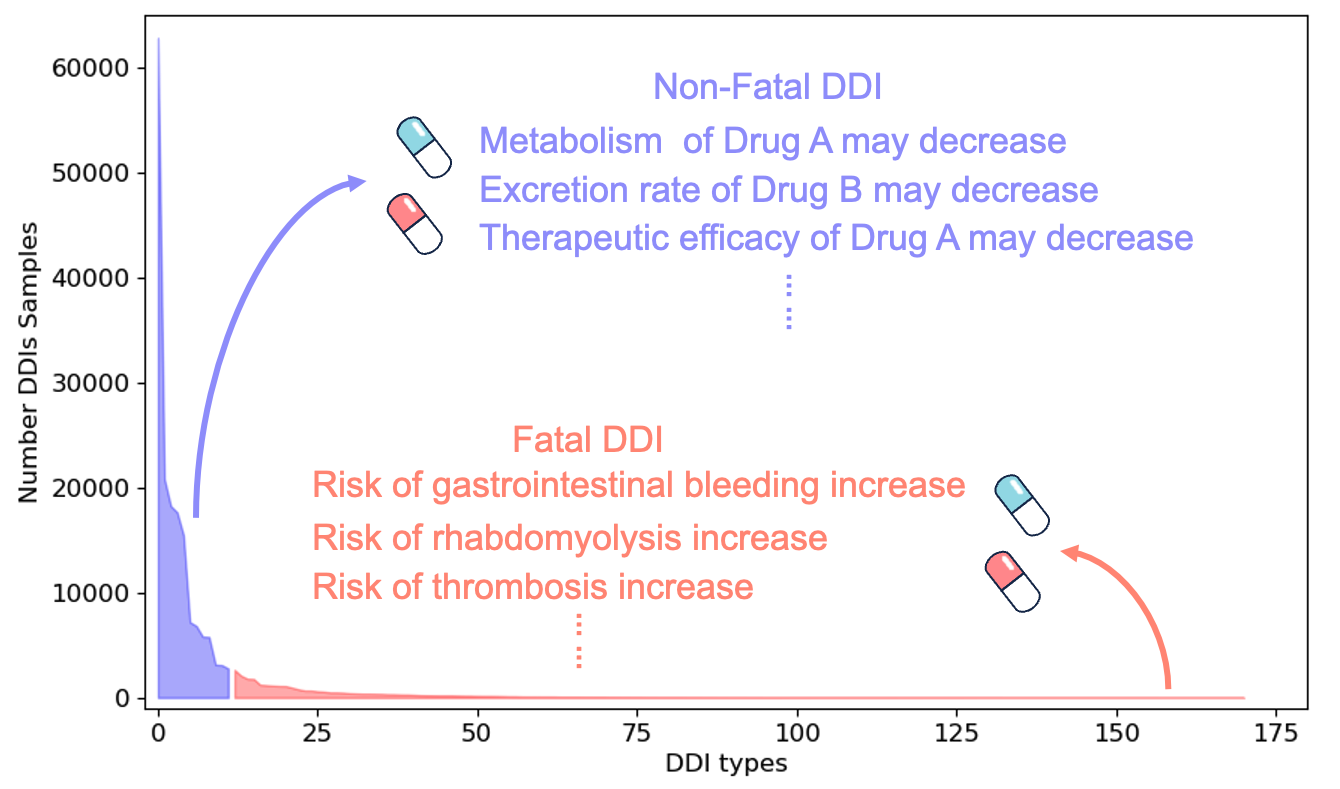}
    \caption{Naturally Long Tailed Distribution of Drug-Drug Interaction in DDI-DB171 dataset}
    \label{fig:DDI_example}
\end{figure}

Recent years have witnessed the emergence of numerous machine learning approaches aimed at enhancing the accuracy and efficacy of DDI prediction. Existing machine learning-based methods have explored the use of Simplified Molecular Input Line Entry System (SMILES) \cite{weininger1988smiles,deng2020multimodal,karim2019drug}, knowledge graph \cite{lin2020kgnn,lyu2021mdnn}, topological similarity \cite{chen2021muffin}, side effects \cite{gottlieb2012indi} for DDIs prediction and achieved a satisfactory result. However, they usually ignore the extremely long-tailed pattern of the distribution of DDIs types, as illustrated in Figure \ref{fig:DDI_example}. The DDI types in the long-tailed region tend to pose greater risks, particularly for patients with chronic illnesses and underlying health conditions. Many of the existing methods are either not designed as multi-classification tasks tailored to handling long-tailed DDI data or are computationally intensive, leading to limited applicability in real-world scenarios. Therefore, the existing methods usually produce a model with high bias to head class, which is dangerous in DDIs prediction task as the tailed DDIs are usually more vital than the head. In addition, although these methods have demonstrated effectiveness to a certain extent, they often overlook the semantic significance of the SMILES system and collaborative interplay between different drug features.

Moreover, long-tailed recognition has been identified as a significant and challenging task in many real-world scenarios such as medical diagnoise and object detection \cite{zhang2023deep,10415681,deathcome,shen2022death,chen2018eeg}. Cost-Sensitive method with a specific-designed loss function is one of the feasible streams to accommodate long-tailed problem. However, many wildly-used loss function such as Focal Loss \cite{Lin_2017_ICCV}, Class Balanced Loss\cite{cui2019class} and Balanced SoftMax \cite{ren2020balanced} only focus on the datasets with class imbalanced rates (CIR) in the range of 50 to 200 and weights the minority classes based on the frequency in the dataset. They employed loss reduction and inverse of class frequency to manipulate the gradient contribution of tail classes. Nonetheless, the gradient reduction mechanisms often result in sub-optimal solutions due to gradient vanishing, and the inverse-frequency strategy tends to produce over-corrected models biased towards tail classes. Furthermore, empirical evidence suggests that existing long-tailed loss functions struggle to generalize effectively on extremely long-tailed datasets, such as those encountered in drug-drug interaction (DDI) prediction, where the imbalance factors can reach 30,000 and beyond.

To address the above limitations, we propose a Tail-Focused Multi-Modal Framework (TFMD) for DDIs task. TFMD not only effectively utilises a range of drug information modalities, including biochemical graphs of drugs, targets, and enzymes, but also introduces molecular spatial structure of drug for prediction using SMILES. Instead solely on the straightforward concatenation of features extracted from distinct modalities, TFMD introduces a modality enhancement fusion approach. This strategy enhances the learning procedure by skillfully amalgamating both the semantic essence of molecule spatial structure representations and the spatial significance of its chemical graph. To maximize the utilization of multimodal information, the distinct information from each modality is late-fused to consolidate the influences of targets and enzymes to drive the ultimate prediction. Leveraging the capabilities of this multi-modal framework. So that TFMD is able to effectively capture the inherent data features and intricate interconnections that exist among various modalities.

Moreover, we draw inspiration from the Focal Loss\cite{Lin_2017_ICCV}, known as a representative re-weighting method in handling imbalanced class data distribution. Traditionally, focal loss tends to diminish the penalty across all classes, inadvertently leading to gradient vanishing and sluggish model convergence in extremely long-tailed data distribution. To address these limitations, we devised a novel Tailed Focal loss (TFL) function, tailored to emphasise the significance of less frequency DDIs types within the long-tail distribution and maintain gradient. In short, the main contributions of this work are summarized as follows.

\begin{itemize}
    \item A novel Tail-Focused Multi-Modal Framework for DDI tasks, the framework adeptly combines various drug information modalities, particularly the molecular spatial structure of drugs, through modality enhancement, effectively capturing complex interconnections among these modalities.
    
    \item A Tailed Focal Loss (TFL) is proposed, the TFL overcomes gradient issues, emphasizing less frequent DDI types within the extremely long-tail distribution, with main contributions outlined.
    
    \item Intensive experiment is conducted to demonstrate the effectiveness of TFMD achieves state-of-the-art performance across four demanding long-tail distributed datasets, significantly surpassing the performance of other SOTA methods.

    \end{itemize}

\section{Related works}
\subsection{Similarity-based Methods}
Similarity-based methods measure the similarity of two drugs by considering the features of drugs such as chemical structure, target, enzyme, and reaction pathway. The researchers who endorse the similarity-based method believe that drugs with similar characteristics are likely to exhibit interactions. DeepDDDI \cite{ryu2018deep} uses Multi-layer Perception (MLP) framework to construct a structural similarity profile (SSP). A binary fingerprint is extracted from chemical structure to learn the internal relations based on the Jaccard similarity of two drugs in terms of SMILES, which is calculated by the number of common chemical fingerprints and the number of union fingerprints of two drugs as shown in Eq. (\ref{eq:Jaccard Similarity}). A and B are bit vectors with binary values, in which 1 represents the presence of a chemical substructure and 0 otherwise. Furthermore, Deng et. al \cite{deng2020multimodal} improved the work by a multi-modal framework, DDIMDL, leveraging not only the SMILES but also the target, enzyme and pathway features as supplement information. DDIMDL simply late-aggregate the votes from modalities without actual fusion can also lead to low performance of model \cite{bayoudh2021survey,xu2024reliable}. MDDI-SCL \cite{lin2022mddi} and MDF-SA-DDI \cite{lin2022mdf} further explore the similarity methods by leveraging supervised contrastive learning and auto-encoder to further improve the performance. However, hashing the SMILES string to the bit binary vectors can potentially abandon the spatial and semantic information presented in SMILES chemical language.

\begin{equation}
    Jaccard\ Similarity = \frac{A \cap B}{(A \cup B) - (A \cap B)}  \label{eq:Jaccard Similarity}
\end{equation}

Similarity-based methods have shown their effectiveness on DDI prediction tasks, and highlighting the importance of chemical structure and other related information such as enzyme and target can improve performance. However, DDIMDL and DeepDDI hashed the molecule spatial structure to a binary bit vector can potentially ignore the informative spatial and semantic information within the SMILES system.

\subsection{Graph-based Methods}
Graph-based Methods consider the knowledge graph and the spatial information of the chemical structure as the main factors causing interaction. The researchers who endorsed graph-based methods believed that the information from knowledge graph of drugs and chemical structure can reveal the potential correlations between atoms and keys by aggregating neighbourhood information to embed the latent features of drugs \cite{lin2020kgnn,wang2022predicting}. KGNN \cite{lin2020kgnn} employed a GraphSAGE architecture to locally aggregate information from 2-Hop neighbours to embed the drugs. LaGAT \cite{hong2022lagat} applied a multi-head graph attention neural networks architecture to learn the attended features around the drugs. DSN-DDI\cite{dsnddi} exploited the intra and inter-view graph embedding method that learns the drug embedding by aggregating information in itself graph and the graph of interacted drugs. However, KGNN, LaGAT, DSN-DDI formulated DDI prediction as a binary classification problem. Even worse, LaGAT was evaluated on a balanced binary dataset which ignored the long-tailed nature of DDIs prediction. MDNN \cite{lyu2021mdnn} and MUFFIN \cite{chen2021muffin} extended the binary classification to a multi-classification task. MUFFIN converted the SMILES to a knowledge graph and bio-medicine graph, and embedded the graphs using the two pre-trained models, TransE \cite{bordes2013translating} and GIN \cite{hu2019strategies} respectively. Specifically, MUFFIN took the cross-product and dot-product of embedding of knowledge graph and bio-structure graph to obtain feature maps, and extract the interaction relationships by a stack of convolutional layers. However, MDNN and MUFFIN failed to address the long-tailed nature of DDI data, and the inappropriate feature fusion lacks interpretation from the perspectives of biology and chemistry.

\subsection{Long-tailed Learning}
Long-tailed Learning is one of the challenging problems in classification, aiming to learn a unbiased deep model from an imbalanced and multi-class classification dataset, also known as long-tailed dataset \cite{zhang2023deep} as shown in Figure. \ref{fig:DDI_example}. It is particularly useful in object detection since only a few objects are the interests of problems. Typical classification loss such as Cross Entropy ignores the imbalance dataset distribution and tends to generate uneven gradient contribution from different classes, leading to a biased model that performs poorly on tailed classes. To address long-tailed problems, one promising approach is to re-weight the gradient contribution from tailed and head samples by a sophisticated loss function. Focal Loss (FL Loss) \cite{Lin_2017_ICCV} and Class Balanced Loss (CB Loss) \cite{cui2019class} employed a similar strategy to reduce loss generated by both head and tailed samples by an adaptive weight term. This weighted term permutes the gradient contribution of tail and head classes by reducing the losses on head samples. However, this approach empirically works well in less imbalanced datasets. We will discuss the reason in Section \ref{section:cost-sensitive learning} and empirically demonstrate the effects of this strategy on different imbalanced rates. In addition, another strategy approaching the long-tailed problem is to weight the loss by the inverse of class frequency. Weighted Cross Entropy Loss (WCE Loss) constructs weighted terms by the inverse of class frequency in the dataset. Balanced Softmax Loss (BS Loss) \cite{ren2020balanced} essentially penalized samples by an additional log inverse of class frequency to the loss. Similarly, LDAM Loss \cite{cao2019learning} also penalized the tail samples by a normalized inverse frequency term. However, this approach often over-emphasizes tailed samples and produces a biased model to tail classes, which loses the generalization ability of head samples with lower overall performance and performs poorly on extremely long-tailed problems.

\begin{table*}[]
    
    \centering
    \scalebox{0.8}{
    \begin{tabular}{c|ccc}
        \hline
         Loss & Standard Formulation & Expand Formulation & Loss of Tail Class \\[3pt]
         \hline
         Cross Entropy & $-log(P_y)$ & - & $-log(P_t)$ \\[2pt]
         Weighted Cross Entropy & $-\frac{\sum^k_{i=1}n_i}{n_y} log(P_y)$ & - & $-\sum^k_{i=1}n_i log(P_t)$ \\[2pt]
         Focal Loss \cite{Lin_2017_ICCV} & $-(1-P_y)^{\gamma} log(P_y)$ & - & $-(1-P_t)^{\gamma} log(P_t)$ \\[2pt]
         Class-balanced Loss \cite{cui2019class} & $-\frac{1-\lambda}{1-\lambda^{n_y}} log(P_y)$ & - & $-\frac{1-\lambda}{1-\lambda^{n_y}} log(P_t)$  \\[2pt]
         Balanced Softmax Loss \cite{ren2020balanced} & $ -log \left( \frac{n_y exp(z_y)}{\sum^k_{i=1} n_i exp(z_i)} \right) $ & $ -\left[ log\left(P_y \right) + log \left(\frac{n_y}{ \sum^k_{i=1} n_i} \right) \right]$  & $ -\left[ log\left( P_t \right) + log \left(\frac{1}{ \sum^k_{i=1} n_i} \right) \right]$   \\[2pt]
         LDAM Loss \cite{cui2019class} & $ -log\left(\frac{exp(z_y - \Delta_y)}{\sum^k_{i=1} exp(z_i - \Delta_i)} \right) $ & $ - \left[ log \left( P_y \right) + log \left( \frac{\sum^k_{i=1}exp(\Delta_i)}{exp(\Delta_y)} \right) \right]$ & $ - \left[ log \left( P_t \right) + log \left( \frac{\sum^k_{i=1}exp(\frac{C}{n_i^{1/4}})}{exp(C)} \right) \right]$  \\[2pt] 
         \hline
    \end{tabular}
    }
    \caption{Summary of Long-Tailed Loss Function Expression. In the table,$P_y$ indicates the probability of ground true label. $P_t$ indicates the probability of tailed classes. $n_y$ is the number of samples in the ground true class. $n_i$ is the number of samples in class $i$. $\lambda$ is the parameter used to determine the effective number in CB Loss, set as 0.999 for the optimized solution suggested by \cite{cui2019class}. $\Delta_i$ is the margin of class $i$, which can be determined by $\frac{C}{n_i^{1/4}}$, where $C$ is a hyper-parameter to be tuned in the range of 0 to 1. $z_i$ is the logits of class $i$. $\gamma$ is a focal parameter set to 2 as suggested by \cite{Lin_2017_ICCV}.} \label{tab:Long-tailed Loss}

\end{table*}

\subsection{SMILES (Simplified Molecular Input Line Entry System)} is a widely used chemical information language with specific grammar to store chemical information based on graph theory for further processing by machine \cite{weininger1988smiles}. SMILES represented the fundamental idea of chemical compounds using a linear sequence of characters. Each atom in chemical compounds can be represented as element symbols and the bounds between each atom are represented by specific characters \cite{sidorova2015bridging,hirohara2018convolutional,weininger1988smiles}. The SMILES notation has specific rules for complicated chemical features such as double bonds, aromatic rings, and functional group, chirality. These rules ensure that the graph properties of the molecule is accurately represented in this molecule spatial structure.

\subsection{Cost-Sensitive method for Long-Tailed Learning} \label{section:cost-sensitive learning}
Cost-Sensitve method is a promising approach to address the long-tailed challenge in many domains such as Computer Vision, Natural Language Processing, and medical diagnoise \cite{zhang2023deep,zhang2024geniu,su2021positive}. It is achieved by a sophistical loss function to re-weight the gradient contribution from head and tail samples. We take Focal Loss as an example to demonstrate the idea of the cost-sensitve method and how it permutes the gradients of tail and head classes by weights. Formally, Focal Loss can be expressed as Eq. \ref{eq:Focal Loss}, where $P_y$ is the true probability of the supervised signal. Focal Loss employs a re-weighting strategy that diminishes the gradients of all classes, with a more substantial reduction for head samples to emphasize the gradient contributions of tail samples. Let's assume a head sample produced high confident $P_y$ e.g. 0.9 due to a large number of samples in the dataset and a tail produced low confident $P_y$ e.g. 0.1 and $\gamma$ is set to 2 as suggested by \cite{Lin_2017_ICCV}. The focal terms $(1-P_y)^{\gamma}$ tend to reduce the loss of the head sample by $0.01$ times while the loss deduction on the tail sample is only $0.81$.

\begin{equation}
    FL = -(1-P_y)^{\gamma} log(P_y) \label{eq:Focal Loss}
\end{equation}

The effectiveness of Focal Loss in managing the long-tailed distribution stems from its capacity to dynamically adjust the impact of each sample based on its predicted probability. This dynamic adjustment ensures that the model does not become overly confident in predictions for head classes and dominate the learning process. By prioritizing the learning of tail samples, Focal Loss helps in achieving a more balanced performance across classes, which is crucial for tasks involving significant class imbalance.

Table \ref{tab:Long-tailed Loss} presents the expressions for Cross Entropy (CE) Loss and five widely-used loss functions for long-tailed learning. Specifically, Focal Loss and Class-Balanced Loss introduce the loss reduction terms $(1-P_y)^{\gamma}$ and $\frac{1-\lambda}{1-\lambda^n_y}$ to the standard Cross Entropy, where these terms are always less than 1 to deduct the loss of all classes and more deduction on head classes. The weight term in Class-Balanced Loss is more pronounced than in Focal Loss because it directly raises $\lambda$ to the power of $n_y$ to achieve greater loss reduction. Nevertheless, this strategy can potentially lead to gradient vanishing, resulting in slow training and suboptimal solutions. Weighted Cross Entropy (WCE Loss), Balanced Softmax Loss (BSM Loss), and LDAM Loss utilize the inverse of class frequency. WCE Loss simply multiplies the inverse of class frequency $\frac{\sum^k_{i=1} n_i}{n_y}$ with the standard CE Loss as a weight to increase the gradient contribution of tail samples. Similarly, BSM Loss implicitly regularizes CE using the logarithm of class frequency $\log(\frac{n_y}{\sum^k_{i=1}n_i})$, as stripped in the Expanded Formulation in Table \ref{tab:Long-tailed Loss}. LDAM Loss also incorporates the logarithm of class frequency $\log(\frac{\sum^k_{i=1}exp(\Delta_i)}{\exp(\Delta_y)})$, where $\Delta_i$ is determined by the inverse of the number of samples in class $n_i$. This approach is particularly challenging to generalize well on datasets with significant discrepancies between head and tail samples, especially on extremely long-tailed datasets with 30,000 classes or more. The frequency of tail samples approaches $\frac{1}{\sum^k_{i=1}n_i}$ in extremely long-tailed datasets. The inverse frequency can generate overwhelming gradients from tail samples, leading to overcorrection, which tends to ignore the significance of head samples. This phenomenon can be empirically demonstrated by extremely high recall and low F1 and precision scores.

\section{Prpoposed Method}
\begin{figure*}[t]
    \centering
    \includegraphics[scale=0.35]{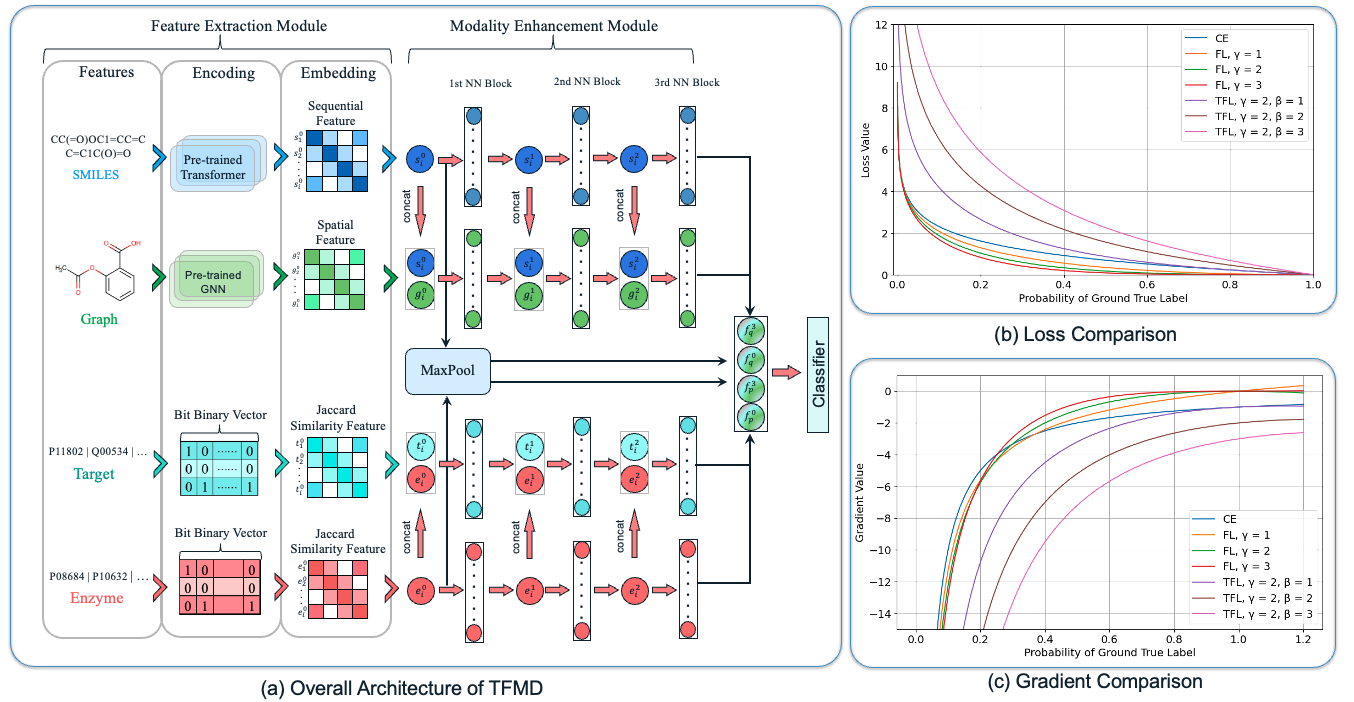}

    \caption{The framework of the proposed method (TFMD): (a) \textbf{Model Architecture}: Graph and Target modalities are considered as strong modality and enhanced by smiles sequential representation and enzyme representation respectively. Enhancement module consists of four independent MLP to extract the feature representation of all the features. The initial modalities are max-pooled to obtain the unique and distinct features of each modality, and then concatenated with fused representation at the end. All the features are concatenated and fed to a 4-layers NN classifier (Right End) for multi-classification prediction depending on the training dataset. (b) \& (c) \textbf{Loss Comparison}: Given a tail class $c_t$, TFL exhibited considerable gradients and higher loss than FL and CE loss. The gradient offset mechanism in TFL recovers gradient back to the level of CE at least and maintain considerable gradient as $P_y \rightarrow 1$ instead of vanishing as shown in (c).} 
    
    \label{fig:overall architecture}
\end{figure*}

\subsection{Problem Definition}
There are potential DDI between drugs $d_i$ in a drug set $D$. We considered DDI prediction as a multi-classification problem, and defined the DDI matrix $M_d$ with shape $|D| \times |D|$ where $|D|$ is the number of drugs. In the DDI matrix, zero indicates no interaction between two drugs and $Y_i$ indicates a specific DDI type. We integrate MILES graph representation $g_i$, SMILES sequential representation $s_i$, target representation $t_i$ and enzyme representation $e_i$ to predict DDIs, denoted by feature matrices $G$, $S$, $T$, $E$ respectively. For each drug $d_i$, there are features $f_i \in [g_i, s_i, t_i, e_i]$.

\subsection{Overview of TFMD}
Tailed Focal Multi-Modal Model for Drug-Drug Interaction (TFMD) is built based on the hypothesis from biomedical perspective, aiming to learn the actual pattern with more effective training. TFMD consists of three parts, modalities enhancement, short-cut fusion and Tailed Focal Loss for long-tailed data distribution. To better understand the design of TFMD, we state the following hypothesis proven by practical biomedical evidences \cite{orr2012mechanism,h2011significance,tokunaga2018understanding}:

\begin{hyp}
    Variations on SMILES chemical structure can potentially produce different, even diametrically opposed, efficacy outcomes.
\end{hyp}

\begin{hyp}
    The drugs sharing the same target and enzyme can potentially produce competition interactions such as inactive behaviours.
\end{hyp}

 TFMD propose a modality enhancement technique that accepts the two modality to enhance the strong modality with weak modality. Specifically, TFMD enhances graph and target modality with SMILES and enzyme respectively. We claimed that the GNN model may fail to capture the actual spatial information about the drugs since the GNN model assumed that the graph is permutation invariant, ignoring the important spatial properties such as chirality presented in SMILE string. Those particular spatial information in molecular science, however, can significantly change the property of a drug and impact the interaction between molecules \cite{nguyen2006chiral}. SMILES symbols can provide the particular 3D spatial information of a drug structure. Therefore, we considered molecule spatial structure as a supporting feature to supply spatial information to graph and enhance the feature representation at each stage. Moreover, many biological molecules, such as protein, enzyme, target and DNA often bind to one specific enantiomer excluding the mirror counterpart molecules resulting in competition interactions such as inactive behaviours \cite{h2011significance}. By enhancing target representation with enzyme modality, we facilitate the modeling of interactions between shared targets and enzymes. This dual enhancement approach is underpinned by the hypothesis that incorporating spatial information and target-enzyme interactions can lead to more informative DDI interaction model. In addition, the latent representations derived from each modality are subjected to a max-pooling operation, which captures the most distinctive and unique fine-grained features of each modality. These features are then concatenated with the enhancement modalities. In addition, inspired by focal loss, we proposed a novel loss function, Tailed Focal Loss (TFL) enabling the model to focus more on tailed sample and address gradient vanishing problem of FL \cite{hossain2021dual} and further boost the performance in extremely long-tailed scenario. TFL adopt an intuitive strategy to offset the gradient reduction of FL, retaining considerable gradients and faster convergence.

\subsection{Modality Fusion}
SMILES string is initially embedded into $g_i$ by pre-trained GNN and $s_i$ by pre-train molecular transformer trained on 100 million SMILES string from ZINC-15 dataset \cite{irwin2022chemformer,qiu2023pre}. Target and enzyme are firstly encoded by binary vector, where 1 indicates presence of a specific target or enzyme. Thus, the embedding of target and enzyme are produced by Jaccard similarity measures with other drugs in the dataset. For each drug $d_i$ with graph $H$, the overall graph representation $g_i$ of $d_i$ can be represented as the aggregation of all the nodes $h_v \in H$. The node representation $h^k_v$ is updated over $k$ iterations by aggregating neighbour nodes $h^{k}_u$ and representation of itself $h^{k-1}_v$ in $k-1$ iteration \cite{hu2019strategies}. The aggregation and message passing process is conducted by multiple weight-share neural networks. Formally, the nodes representation after $k$ iterations is: 

\begin{equation}
    h^{k}_v = \sigma(h^{k-1}_v, AGG(h^{k-1}_u |\ \forall u \in N(v)))
\end{equation}
where $\sigma$ is the non-linearity function (e.g. ReLU), $N(v)$ is the number of neighbors of node $u$, $AGG$ is the aggregation function. 

To obtain the overall representation of drug $d_i$, a READOUT function is applied to pool the nodes features $h^{k}_u$ such as average pooling and max pooling\cite{you2019position,hu2019strategies}

\begin{equation}
    g_i = READOUT(h^k_v |\ \forall v \in H)
\end{equation}

Before each fully connected block in fusion module, we concatenated the corresponding $g^k_i$ with $s^k_i$ to fuse the particular spatial information including chirality, functional groups to $g^k_i$ and $t^{k}_i$ with $s^{k}_i$ to enhance the representation in latent space. Finally, the initial embeddings of each modality $g^{0}_i$, $s^{0}_i$, $t^{0}_i$, $e^{0}_i$ are pooled by a max-pooling layer and concatenated to the fused representation, producing final embedding $F_u$. Therefore, two drugs $d_q$ and $d_p$ producing potential interaction are concatenated to form the interaction representation.

\begin{algorithm}
\caption{TFMD Algorithm}
\label{algm:TFMD algoritm}
\begin{algorithmic}[1]
\STATE  \textbf{Input:} Available drug features $f_i \in [g_i, s_i, t_i, e_i]$ and DDI Matrix $M_d$ with drug $d_i$
\STATE \textbf{Output:} Predicted DDI types $Y_i$ for drug-pairs $(d_p, d_q)$
\WHILE {TFMD not Converge}
    \FOR{$d_i$ in $D$}
        \STATE $g^k_i \gets \sigma (CONCAT(g^{k-1}_i, s^{k-1}_i))$
        \STATE $t^k_i \gets \sigma (CONCAT(t^{k-1}_i, e^{k-1}_i))$
        \STATE $f^k_i \gets \sigma(f^{k-1}_i)$
    \ENDFOR
    \STATE $f^0_i \gets MaxPool(f^0_i)$
    \STATE $F_u \gets CONCAT(g^k_i, s^k_i, t^k_i, e^k_i, g^0_i, s^0_i, t^0_i, e^0_i)$
    \FOR {$(d_q, d_p) \in M_d$}
        \STATE $P_y(Y_i|F_u) \gets MLP(d_q, d_p)$
        \STATE $loss \gets TFL(P_y(Y_i|F_u), Y_i)$
    \ENDFOR
\ENDWHILE
\end{algorithmic}
\end{algorithm}

\subsection{Tailed Focal Loss}
In extremely long-tailed setting, the number of tailed class $N_t$ is dramatically lower than the number of head class $N_h$, $N_t \ll N_h$, and head classes dominate the dataset with large number of samples $N_h \cong N$. Therefore, the frequency of the tail classes becomes $\frac{1}{\sum^k_{i=1} n_i}$. Consequently, this allows for the simplification of existing loss functions in extremely long-tailed setting and obtain the loss for tail class as shown in Table. \ref{tab:Long-tailed Loss}.

For WCE Loss, the loss of the tail class will be dependent on the number of samples in the dataset. Similar to BS Loss and LDAM Loss, the loss of tail class adds a term that depends on the sample size of the dataset to standard CE loss. Empirical evidence in extremely long-tailed settings suggests that this approach can lead to overcorrected models characterized by high recall but low precision due to the substantial imbalance factors. LDAM attempts to alleviate this overcorrection by introducing square roots and softmax functions to mitigate the effects of large sample sizes. However, even this approach often results in suboptimal performance in extreme long-tailed scenarios. Focal Loss and Class-balanced Loss, on the other hand, stand for loss reduction strategy to reverse the gradient contribution of head and tail classes. Focal Loss maintains a consistent loss expression for both head and tail classes, utilizing the focusing parameter to attenuate the loss based on the confidence $P_y$. The upper bound of Class-balanced Loss is cross entropy. Unlike focal loss, CB loss depends on the number of samples in classes to deduct loss usually generates a worse gradient vanishing compared to Focal Loss, leading gradient vanishing during backpropagation and resulting in exceedingly slow training due to the gradient reduction mechanism \cite{hossain2021dual,li2019dual}.

To this end, we proposed a novel loss function, Tailed Focal Loss (TFL), that adds an additional loss term, $log(P_y^{\beta F_{T_s}(c_y)})$, to avoid gradient vanishing by offsetting the gradient reduction of tailed samples in FL, where $P_y$ is the predictive probability of ground-true label, $F_{T_s}(c_y)$ is a threshold function that determines whether or not the input class is a tailed class $c_y$ and $\beta$ is the hyper-parameter to emphasize how strong the penalty applied to tailed classes. The threshold function $F_{T_s}(c_y)$ returns 1 if the corresponding class $c_y$ is a tailed class, and 0 otherwise. The recommended value for $\beta$ is 2 given by experiments. For those head class, we calculated the loss as usual focal loss with $\gamma = 2$. The impacts of $F_{T_s}(c_y)$ and $\beta$ will be discussed by incremental experiments in the experiment section.

In Figure \ref{fig:overall architecture} (b) and (c), we depict the loss and gradient plots of FL, TFL, and their variants to elucidate the impact of $\beta$ and $\gamma$, alongside the gradient loss phenomenon. It becomes evident that the gradients and penalty begin to decline markedly while $P_y$ remains small in FL, leading to under-confidence and sluggish training. Conversely, Tailed Focal Loss (TFL) initiates with more substantial gradients and maintains substantial gradients even as $P_y$ gains confidence. This dynamic approach effectively prevents gradient vanishing and allocates greater focus to tailed classes. We augment the gradient contribution from tailed classes through an additional term tailored to these specific classes. Formally, the proposed TFL loss function is defined as follows:

\begin{equation}
    TFL = -[(1-P_y)^\gamma log(P_y) + log(P^{\beta F_{T_s}(c_y)}_y)]    \label{eq:TFL}
\end{equation}

To determine whether or not a class $c_t$ belongs to a tailed class, all classes are sorted based on their corresponding number of samples $|c_i|$ in descending order, and the sample normalized position $P_{c_y}$ is determined by the ratio of the cumulative summation of samples from the head class to the current class $\sum_{i=0}^{M} |c_i|$ and the total number of DDIs $\sum_{j=0}^{N_c} |c_j|$, where $M$ is the index of current class in the sorted classes rank, $N_c$ represents the number of DDI classes in dataset. If the normalized position of the class $c_y$ surpasses a specified threshold $T_s$, (e.g., 0.9), the class is deemed to belong to the tailed group. $T_s$ decided the number of tail class $N_t$ that will be emphasized, the greater $T_s$ is, the less tail class will be. Meanwhile, a higher $\beta$ generate higher gradient and strong adjustment of gradient proportion from tailed class. We considered $T_s$ as a hyper-parameter providing the best performance at 0.9 in our experiments across all datasets.

\begin{equation}
    P_{c_y} = \frac{\sum_{i=0}^{M} |c_i|}{\sum_{j=0}^{N} |c_j|} \label{eq:sample normalization},  \ \ \ \  F_{T_s}(c_y)=\begin{cases}
    1, & \text{if $P_{c_y}>T_s$}.\\
    0, & \text{otherwise}.
  \end{cases}
\end{equation}

We claim that FL cannot retain sufficient adjustment to the proportion of gradient amount in long-tailed scenario due to gradient vanishing. TFL, on the other hand, addresses the problem by introducing a tail regularizer. Intuitively, the additional regularizer in TFL specifically "double" penalize the tailed samples and offer a supplement of gradient to tailed classes. Unlike the gradient reduction mechanism in FL, TFL offsets the gradient reduction of tailed classes. In the following session, we show the gradient upper bounds of FL and TFL to compare the analytic solution of threshold $P_y$ where the gradient vanishing occurs. Eq. (\ref{eq:gradient FL}) and (\ref{eq:gradient TFL}) are the derivatives of FL and TFL given a tailed sample. It is obvious that the FL reduce the gradients by $(1-P_y)^{\gamma - 1}$ for any input $P_y > 0$ leading to an upper bound gradient shown in Eq. (\ref{eq:upper bound of gradient FL}).

\begin{figure}[h]
    \centering
    \begin{subfigure}[b]{0.42\textwidth}
        \centering 
        \includegraphics[width=\textwidth]{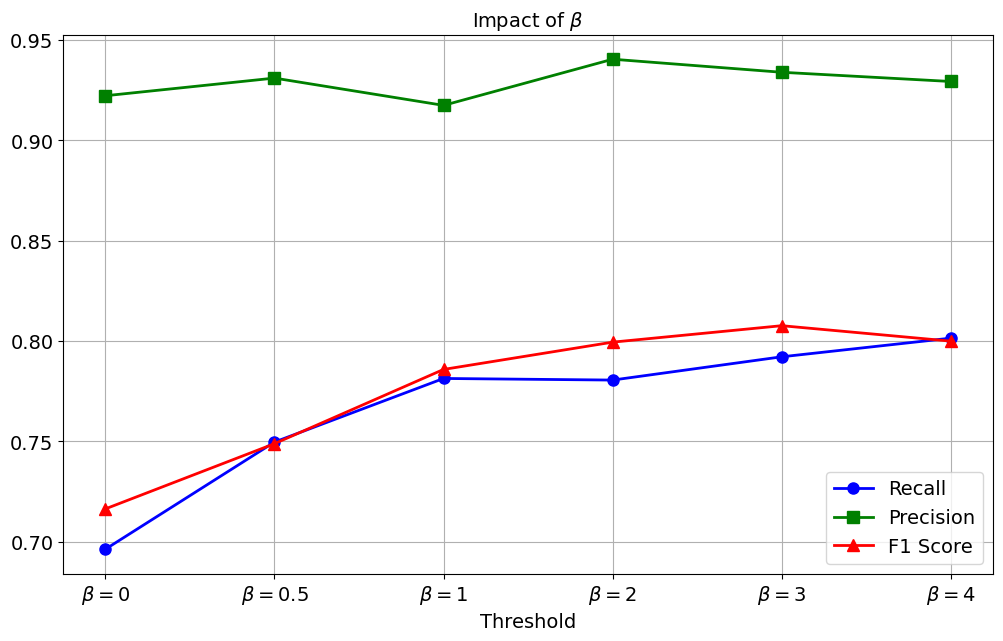}
        \caption{ Incremental Experiment on $\beta$}
        \label{fig:Incremental Experiment on beta}
    \end{subfigure}
    \hfill
    \begin{subfigure}[b]{0.42\textwidth}
        \centering
        \includegraphics[width=\textwidth]{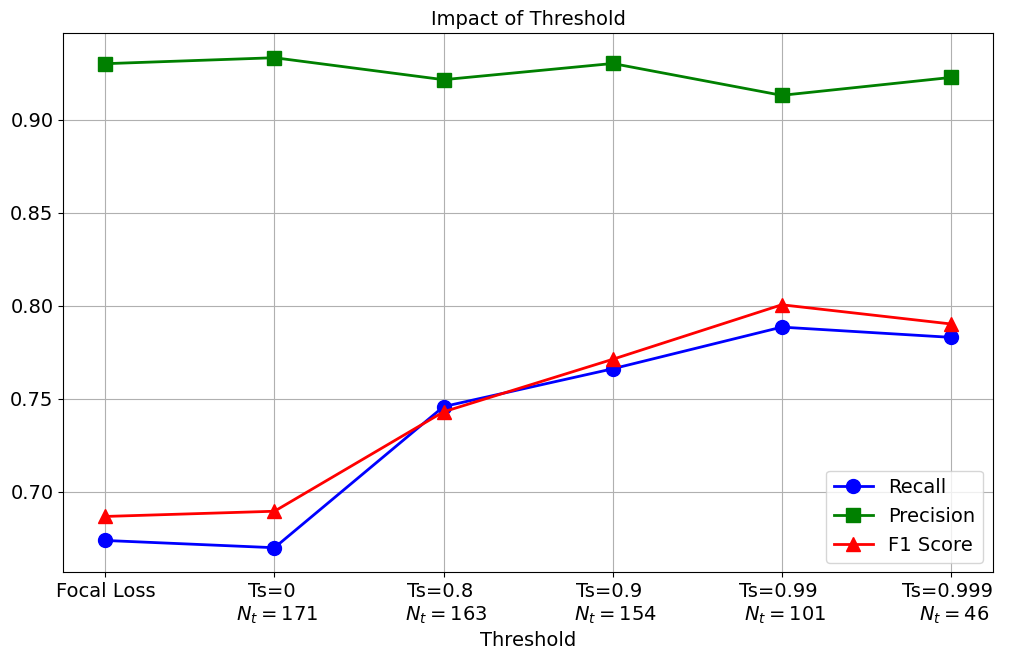}
        \caption{Incremental Experiment on $T_s$}
        \label{fig:Incremental Experiment on Ts}
    \end{subfigure}
    \caption{Incremental Experiments for Parameter Tuning}
    \label{fig:experimental graphs}
\end{figure}

\begin{table*}[hbt!]
    \small
    \centering
    \caption{Performance Comparison over DDI-DB110 and DDI-DB171}
    \setlength{\arrayrulewidth}{0.4mm}
    \renewcommand{\arraystretch}{1.2}
    \label{table:performance comparison DDI-DB110 and DDI-DB171}
    \resizebox{16cm}{1.8cm}{%
    \begin{tabular}{c|cccccc|cccccc}
        \hline
        \multirow{2}{*}{\centering \textbf{Method}} &\multicolumn{6}{c|}{DDI-DB110}& \multicolumn{6}{c}{DDI-DB171}\\

           & Accuracy & Precision & Recall & F1 & AUC & AUPR & Accuracy & Precision & Recall & F1 & AUC & AUPR \\
        \hline
        SVM & 0.3306 & 0.8042 & 0.0213 & 0.0955 & 0.8080 & 0.0835 & 0.3463 & 0.8708 & 0.0202 & 0.0476 & 0.7381 & 0.0938 \\
        DeepDDI \cite{ryu2018deep}& 0.6022 & 0.6092 & 0.2105 & 0.3540 & 0.9208 & 0.2739 & 0.5919 & 0.7493  & 0.1735 & 0.2532 & 0.8873 & 0.2390 \\
        DDIMDL \cite{deng2020multimodal} & 0.5513 & 0.5418 & 0.2463 & 0.3031 & 0.9801 & 0.5239 & 0.5606 & 0.3708 & 0.1759 & 0.2174 & 0.9664 & 0.4599 \\
        MUFFIN \cite{chen2021muffin} & 0.8661 & 0.7246 & 0.5674 & 0.6109 & 0.9945 & 0.7987 & 0.8807 & 0.5548 & 0.5307 & 0.4833 & 0.9844 & 0.6617 \\
        MDF-SA-DDI \cite{lin2022mdf} & 0.8415 & 0.8512 & 0.7124 & 0.7627 & 0.9771 & 0.8246 & 0.8903 & 0.9115 & 0.5663 & 0.6195 & 0.9913 & 0.8208 \\
        DDI-SCL \cite{lin2022mddi} & 0.9201 & 0.8696 & 0.8784 & 0.8690 & \textbf{0.9976} & 0.9199 & 0.5843 & 0.7123 & 0.6206 & 0.5860 & 0.9446 & 0.6665 \\
        TFMD & \textbf{0.9439} & \textbf{0.9287} & \textbf{0.8919} & \textbf{0.9028} & 0.9970 & \textbf{0.9366} & \textbf{0.9410} & \textbf{0.9205} & \textbf{0.7806} & \textbf{0.7995} & \textbf{0.9959} & \textbf{0.8941} \\
        \hline
    \end{tabular}%
    }
\end{table*}

\begin{table*}[hbt!]
    \caption{Performance Comparison over DDIMDL and MUFFIN}
    \small
    \setlength{\arrayrulewidth}{0.4mm}
    \centering
    \renewcommand{\arraystretch}{1.2}
    \label{table:performance comparison DDIMDL and MUFFIN}
    \resizebox{16cm}{1.8cm}{%
    \begin{tabular}{c|cccccc|cccccc}
        \hline
        \multirow{2}{*}{\centering \textbf{Method}} & \multicolumn{6}{c|}{DDIMDL} & \multicolumn{6}{c}{MUFFIN}\\
    
           & Accuracy & Precision & Recall & F1 & AUC & AUPR & Accuracy & Precision & Recall & F1 & AUC & AUPR \\
        \hline
        SVM & 0.3761 & 0.8496 & 0.0305 & 0.0912 & 0.7921 & 0.0699 & 0.2808 & 0.8146 & 0.0177 & 0.0907 & 0.6921 & 0.0485  \\
        DeepDDI \cite{ryu2018deep}& 0.8457 & 0.7624 & 0.6037 & 0.6493 & 0.9837 & 0.6656 & 0.8974 & 0.8715  & 0.7807 & 0.8016 & 0.9914 & 0.8420 \\
        DDIMDL\cite{deng2020multimodal} & 0.8173 & 0.7935 & 0.5919 & 0.6596 & 0.9864 & 0.7511 & 0.8583 & 0.8595 & 0.7305 & 0.7384 & 0.9969 & 0.8729 \\
        MUFFIN \cite{chen2021muffin} & 0.8258 & 0.6111 & 0.5310 & 0.5498 & 0.9954 & 0.7305 & 0.9360 & 0.8970 & 0.8035 & 0.8351 & \textbf{0.9989} & 0.9296 \\
        MDF-SA-DDI \cite{lin2022mdf} & 0.9109 & 0.8812 & 0.8460 & 0.8263 & 0.9756 & 0.8967 & 0.9408 & 0.9506 & 0.9071 & 0.9353 & 0.9982 & 0.9584 \\
        DDI-SCL \cite{lin2022mddi} & 0.9019 & 0.8333 & 0.8216 & 0.8205 & \textbf{0.9974} & 0.8831 & 0.9553 & 0.9414 & 0.9411 & 0.9391 & 0.9976 & 0.9651 \\
        TFMD & \textbf{0.9190} & \textbf{0.8933} & \textbf{0.8659} & \textbf{0.8486} & 0.9906 & \textbf{0.9013} & \textbf{0.9728} & \textbf{0.9648} & \textbf{0.9455} & \textbf{0.9458} & 0.9949 & \textbf{0.9666} \\
        \hline
    \end{tabular}%
    }
\end{table*}

Such that, we can find the upper bound of $P_y$ where the gradient vanishing starts in Focal Loss by equalizing the gradient of FL to the gradient of CE in Eq. \ref{eq:FL P_y}. If $\lambda = 2$ in practice, gradient contribution from tailed samples of FL will be less than CE at $P_y = 0.61$, higher $\lambda$ leads to worse gradient vanishing:

Tailed Focal Loss, on the other hand, avoids the gradient vanishing problem by offsetting the gradient reduction. To simplify notation, we consider only the gradient of the tailed classes, such that $F_{T}(c_y) = 1$. Similarly, we obtain the gradient upper bound of TFL as shown in Eq. \ref{eq:gradient upper bound TFL}.

\begin{align}
    \frac{\partial FL}{\partial P_y} 
    &= \gamma(1-P_y)^{\gamma-1} log(P_y) - (1-P_y)^{\gamma}\frac{1}{P_y}  \\
    &= (1-P_y)^{\gamma-1} \left[ \gamma log(P_y) - \frac{1}{P_y} + 1 \right] \label{eq:gradient FL} \\ 
    &\le  \gamma log(P_y) - \frac{1}{P_y} + 1  \label{eq:upper bound of gradient FL}
\end{align}

\begin{equation}
     -\frac{1}{P_y} = \gamma log(P_y) - \frac{1}{P_y} + 1 	\Rightarrow  P_y = e^{-\frac{1}{\gamma}} \label{eq:FL P_y}
\end{equation}

\begin{align}
    \frac{\partial TFL}{\partial P_y}  \label{eq:gradient TFL}
    &= (1-P_y)^{\gamma-1} \left[ \gamma log(P_y) - \frac{1}{P_y} + 1 \right] - \frac{\beta}{P_y} \\  
    &\le \gamma log(P_y) - \frac{1}{P_y} + 1 - \frac{\beta}{P_y} \label{eq:gradient upper bound TFL}
\end{align}

\begin{align}
    \gamma \log(P_y) - \frac{1}{P_y} + 1 - \frac{\beta}{P_y} = -\frac{1}{P_y} \Rightarrow \gamma \log(P_y) = \frac{\beta - P_y}{P_y} \\ 
    \Rightarrow P_y = e^{\frac{\beta - P_y}{\gamma P_y}} = e^{\frac{\beta}{\gamma P_y}} / e^{\frac{1}{\gamma}} \Rightarrow \frac{\beta}{\gamma}\frac{1}{P_y}e^{\frac{\beta}{\gamma}\frac{1}{P_y}} = \frac{\beta}{\gamma}e^{\frac{1}{\gamma}}
\end{align}

To compare with FL, we also determine $P_y$ shown in E. \ref{eq:TFL P_y}, where the gradient vanishing starts by equalizing the graident of CE.  We employ Lambert $\mathcal{W}$ Function \cite{corless1996lambert} to isolate $P_y$, where $\mathcal{W}(xe^x) = x$. Therefore, the intersection of TFL and CE is equal to $P_y = 1$ given $\beta=1$ and $\gamma=2$ same as FL, which ensures that the gradient vanishing is eliminated in the range of $P_y \in [0, 1]$. Higher $\beta$ leads to a higher gradient offset from tailed sample. We obtain the optimal solution by setting $\beta=3$. TFL will not lead to overcorrection problem as WCE loss and LDAM loss since the offset gradient is only from the minority class and controlled by $\beta$.

\begin{align}
    \frac{\beta}{\gamma}\frac{1}{P_y}e^{\frac{\beta}{\gamma}\frac{1}{P_y}} = \frac{\beta}{\gamma}e^{\frac{1}{\gamma}} \Rightarrow \mathcal{W}(\frac{\beta}{\gamma}\frac{1}{P_y}e^{\frac{\beta}{\gamma}\frac{1}{P_y}}) = \mathcal{W}(\frac{\beta}{\gamma}e^{\frac{1}{\gamma}}) \\ 
    \Rightarrow \frac{\beta}{\gamma}\frac{1}{P_y} = \mathcal{W}(\frac{\beta}{\gamma}e^{\frac{1}{\gamma}}) \Rightarrow  P_y = \frac{\beta}{\gamma \mathcal{W}(\frac{\beta}{\gamma}e^{\frac{1}{\gamma}})} \label{eq:TFL P_y}
\end{align}

\section{Experiments}
\subsection{Dataset}
Our experimental dataset originates from the publicly accessible benchmark repository DrugBank (V5.1.10) \cite{wishart2018drugbank}, encompassing 4,336 approved drugs. From this pool, we specifically chose 1,178 drugs possessing relevant features, covering a spectrum of 199,052 drug-drug interactions (DDIs) spanning 171 DDI types. The selected drugs are used to form two extremely long-tailed datasets, DDI-DB110, DDI-DB171 with 110 and 171 DDI classes respectively. The DDI-DB110 is the subsets of DDI-DB171, sampling from first 110 DDI types where the DDI types are ranked based on the number of DDIs in descending order. It is worthy to note that the DDI-DB171 dataset is an extremely challenging dataset to train since there are 44 classes are less than 10 samples, and 14 classes have only 2 samples. We employed Class Imbalance Ratio (CIR) defined as the ratio between the number of majority class and the number of minority class to describe the degree of imbalance. We also applied our method to two baseline datasets, DDIMDL \cite{deng2020multimodal} and MUFFIN \cite{chen2021muffin}, in which two state-of-the-art methods were evaluated on. In addition, the number of drugs in DDIMDL is factually not aligned with the original paper since we only considered drugs producing interaction in DDIMDL dataset and removed one drug: DB00515, due to incompatibility issues with the SMILE string and the chemical graph structure production libraries dgl and dgllife. Finally, DDIMDL dataset yields 569 drugs in our experiments. Table. \ref{tab:dataset_table} shows the statistics of the four datasets.

\begin{table}[h]
    \centering
    \captionsetup{skip=5pt}
    \setlength{\arrayrulewidth}{0.4mm}
    \renewcommand{\arraystretch}{1.2}
    \caption{Statistics of the four datasets for method evaluation}
    \label{tab:dataset_table}
    \resizebox{6.6cm}{1.1cm}{%
    \begin{tabular}{c|ccccc}
        \hline
        Name & DDI & Class & Drug & Feature & CIR\\
        \hline
        DDIMDL & 37243 & 65 & 569 & GSTEP & 3270\\
        MUFFIN & 172426 & 81 & 1569 & GS & 5243\\
        DDI-DB110 & 198631 & 110 & 1178 & GSTE & 3304\\
        DDI-DB171 & 199052 & 171 & 1178 & GSTE & 31390\\
        \hline
    \end{tabular}%
    }
    
    \footnotesize{GSTEP in Feature indicates G(Graph), S(SMILES), T(Target), E(Enzyme), P(Pathway)} 
\end{table}

\begin{table*}[h]
    \caption{Performance Comparison of Loss Function}
    \centering
    \label{table:Loss Comparison}
    \resizebox{15cm}{1.55cm}{%
    \begin{tabular}{c|cccccc|cccccc}
        \hline
        \multirow{2}{*}{\centering \textbf{Loss}} & \multicolumn{6}{c|}{DDI-DB110} & \multicolumn{6}{c}{DDI-DB171}\\
           & Accuracy & Precision & Recall & F1 & AUC & AUPR & Accuracy & Precision & Recall & F1 & AUC & AUPR \\
        \hline
        CE & \textbf{0.9473} & 0.9202 & 0.8835 & 0.8998 & 0.9970 & 0.9343 & 0.9353 & 0.9150 & 0.4752 & 0.4886 & 0.9798 & 0.7215  \\
        WCE & 0.8762 & 0.7957 & \textbf{0.9318} & 0.8458 & 0.9962 & 0.9349 & 0.6604 & 0.7061 & \textbf{0.8574} & 0.7534 & 0.9818 & 0.8367  \\
        FL\cite{Lin_2017_ICCV} & 0.9436 & 0.9372 & 0.8841 & 0.8990 & 0.9968 & 0.9332 & 0.9311 & 0.9214 & 0.7028 & 0.7334 & 0.9901 & 0.8458  \\
        CB\cite{cui2019class} & 0.9205 & 0.8669 & 0.9181 & 0.8838 & 0.9961 & 0.9406 & 0.3535 & 0.4467 & 0.7302 & 0.4448 & 0.9411 & 0.7058  \\
        BS\cite{ren2020balanced}  & 0.9466 & 0.8480 & 0.9255 & 0.8760 & 0.9969 & 0.9342 & 0.9316 & 0.6249 & 0.8143 & 0.7693 & 0.9882 & 0.8215  \\
        LDAM\cite{cao2019learning} & 0.9456 & \textbf{0.9422} & 0.8734 & 0.8901 & 0.9918 & 0.8842 & 0.9396 & 0.8996 & 0.7156 & 0.7434 & 0.9719 & 0.6724  \\
        TFL & 0.9439 & 0.9287 & 0.8919 & \textbf{0.9028} & \textbf{0.9970} & \textbf{0.9366} & \textbf{0.9410} & \textbf{0.9205} & 0.7806 & \textbf{0.7995} & \textbf{0.9959} & \textbf{0.8941} \\
        \hline
    \end{tabular}%
    }
\end{table*}

\begin{table}[h]
    \caption{Performance Comparison of TFMD's Variants}
    \centering
    \setlength{\arrayrulewidth}{0.4mm}
    \renewcommand{\arraystretch}{1.2}
    \label{table:Abalation study}
    \resizebox{8.2cm}{1.85cm}{%
    \begin{tabular}{c|cccccc}
        \hline
        \multirow{2}{*}{\centering \textbf{TFMD}} & \multicolumn{6}{c}{DDI-DB171}\\
           & Accuracy & Precision & Recall & F1 & AUC & AUPR \\
        \hline
        TFL-G & 0.8684 & 0.8378 & 0.6926 & 0.6762 & 0.9806 & 0.7869  \\
        TFL-S & 0.8190 & 0.8221 & 0.6085 & 0.6066 & 0.9845 & 0.6775  \\
        TFL-T & 0.8021 & 0.8341 & 0.6518 & 0.6474 & 0.9857 & 0.7822 \\
        TFL-E & 0.5986 & 0.6232 & 0.3220 & 0.3604 & 0.9616 & 0.3677 \\
        TFL-GS & 0.8775 & 0.8801 & 0.6959 & 0.7003 & 0.9809 & 0.7863 \\
        TFL-TE & 0.8851 & 0.8717 & 0.6941 & 0.6871 & 0.9869 & 0.7847 \\
        TFL-GSTE & \textbf{0.9410} & \textbf{0.9405} & \textbf{0.7806} & \textbf{0.7995} & \textbf{0.9959} & \textbf{0.8941} \\
        \hline
    \end{tabular}%
    }
\end{table}

\subsection{Evaluation Metrics}
The DDI prediction is considered as a multi-classification problem in a 'one vs rest' manner, which performs the classification per class as a binary classification. To fairly evaluate the overall performance of the compared methods, Accuracy, Precision, Recall, F1, Area Under the ROC (AUC) and Area Under Precision-Recall Curve (AUPR) are selected to evaluate not only the performance in head-samples, but also in the tail-samples. Macro averaging is used to report the overall performance of all classes. AUPR is more sensitive to imbalanced data than AUC.

\subsection{Baseline Model}
\textbf{DDI Baselines:} We compare 1 traditional statistical baseline, SVM and 5 representative SOTA models, DeepDDI\cite{ryu2018deep}, DDIMDL\cite{deng2020multimodal}, MUFFIN\cite{chen2021muffin}, MDF-SA-DDI\cite{lin2022mdf}, DDI-SCL\cite{lin2022mddi}. DDIMDL is multimodal model integrates with different source of information of drugs with simple late fusion. MUFFIN exploited bimodal fusion with CNN as feature extraction. DDI-SA-DDI utilized auto-encoder and multi-head attention mechanism to fuse DDI vectors. DDI-SCL is a contrastive learning model in the context of DDI classification. Besides, DDI-SA-DDI and DDI-SCL are driven by focal loss to relieve imbalanced data distribution. MUFFIN, MDF-SA-DDI and DDI-SCL have demonstrated their effectiveness and outperformed other methods such as KGNN \cite{lin2020kgnn}. Some DDI approach such as MDNN \cite{lyu2021mdnn} do not release their publicly available repository for reproduction purpose. For this reason, we do not compare all baseline models.

\textbf{Loss Function Baselinses:} We select cross entropy and 5 widely used long-tailed learning loss function from computer vision: Weighted Cross Entropy (WCE), Focal Loss (FL Loss) \cite{Lin_2017_ICCV}, Class-balanced Loss (CB Loss) \cite{cui2019class}, Balanced Softmax Loss (BS loss) \cite{ren2020balanced}, LDAM Loss \cite{cao2019learning}. FL Loss and Class-balanced Loss employed gradient reduction to manipulate the gradient contribution from head and tail samples. WCE Loss, BS Loss and LDAM Loss employed the information from sample frequency to permute the penalty of tail and head.

\section{Results Discussion} 
\subsection{Performance Comparison}

The experiment results in Table \ref{table:performance comparison DDI-DB110 and DDI-DB171} and \ref{table:performance comparison DDIMDL and MUFFIN} indicated that TFMD significantly outperform other baseline models over all the evaluation metrics. The traditional methods such as SVM and DDIMDL failed to handle the tailed class due to the disillusionary recall, F1 and AUPR score. MUFFIN, MDF-SA-DDI, DDI-SCL have expressed relatively good performance on DDIMDL, MUFFIN, and DDI-DB110 datasets. It is worthy noticing that although MDF-SA-DDI and DDI-SCL are also empowered by focal loss with consistent $\gamma$ value, TFMD still outperforms all baselines on all the dataset, indicating the effectiveness and robustness of our modality fusion design to long-tailed data distribution. Although TFMD lose the AUC on DDIMDL, MUFFIN and DDI-DB110 datset, we still demonstrate significant improvement and the best generalization on other metrics. Moreover, TFMD further boosts the overall performance and dramatically improve F1 score to 0.7995 in DDI-DB171, while other baselines are entirely failed due to their vulnerability to long-tailed issues.

\subsection{Effectiveness of Tailed Focal Loss}
As depicted in Eq. \ref{eq:gradient TFL}, the additional loss term theoretically boosts the gradient of tailed classes by addressing the gradient reduction mechanism and gradient vanishing problem in standard FL. We compared different loss functions for long-tailed learning in Table. \ref{table:Loss Comparison}. Incremental experiments shown in Fig. \ref{fig:experimental graphs} indicate that a higher value of $\beta$ leads to increased gradients but better performance, posing a risk of gradient explosion if $\gamma$ is excessively elevated. While a higher threshold $T_s$ limits the number of tailed classes addressed by TFL. Conversely, a lower $T_s$ implies a broader range of classes being emphasized by TFL. It's important to note that setting $T_s = 0$ essentially equates the loss to cross-entropy, whereas not classifying any class as 'tailed' (with $T_s = 1$) reverts TFL to the standard focal loss. In DDI-DB 110, TFL may not be the strongest loss on Accuracy, Precision and Recall to other loss functions, but the most generalized ability as indicated by the highest F1, AUC and AUPR scores. TFL demonstrate the most balanced Recall-Precision trade-off and highest AUPR, indicating our loss function achieve the optimal solution to balanced head and tail samples with least bias compared to other loss. In addition, we implement the experimental setting with Cross Entropy, the performance is relatively good even outperforming CB Loss and WCE Loss settings, indicating that our performance not only relys on the TFL loss but also the comprehensive designs of our model framework. In DDI-DB171 TFL demonstrates significant improvement over all long-tailed loss, the most challenging dataset in our experimental setting. WCE Loss, CB Loss, BS Loss over-emphasize the tail samples and lose the generalization to head samples. We can find the primary reason from the formulation of the loss shown in Table. \ref{tab:Long-tailed Loss}. The inverse frequency strategy failed to consider the model ability without logits information. 
Those loss functions usually over-correct the importance of head sample and is sensitive to the large discrepancy between the number of head and tail classes. Although FL Loss and LDAM Loss demonstrate the most competitive performance, the gradient reduction mechanism hinders FL Loss from the optimal solution due to serious gradient vanishing. LDAM, on the other hand, relief the over-correction problem by introducing a softmax with the temperature that smooths the gradient. It is still difficult to balance the head and tail samples from the results of DDI-DB171.

\subsection{Ablation Study}
In real-world scenarios, ensuring the availability of all features is challenging. To investigate the impacts of different features and their combinations, we compared the performance of TFMD and its six variants, as presented in Table \ref{table:Abalation study}. Notably, the graph variant and the target variant achieved the highest F1 scores among the single-feature variants, with scores of 0.6762 and 0.6474, respectively, aligning with our prior assumptions and intuition. The GS variant, which incorporates both graph and spatial information, demonstrated a 3\% improvement in F1 score over the graph-only variant. This result highlights the supportive effectiveness of spatial information in molecular string representations. Although the enzyme feature alone proved to be relatively weak, its combination with the target feature revealed significant interactions, enhancing the performance of the single-target variant by 4\% in both F1 score and recall. The consistent superior performance of multi-feature variants over individual features underscores the necessity of each component in achieving optimal results. These findings suggest that leveraging multiple features in combination provides a more comprehensive understanding and improved predictive capability compared to using any single feature in isolation.

\section{Conclusion}
Overall, the proposed method addresses the long-tailed distribution and gradients vanishing issues in extremely long-tailed dataset of focal loss using a novel loss function and outperform the existing loss function adapted from computer vision. Our proposed architecture outperforms over all baseline models in four challenging datasets. Our model architecture is more comprehensive and interpretable than other methods from chemistry perspective, as it correctly leveraged the graph and particular spatial information in SMILES system, Target, Enzyme of drugs to construct a strongly fused feature set for decision making.

\section{Acknowledgement}
This work was supported by Australian Research Council Linkage (Grant No. LP230200821), Australian Research Council Discovery Projects (Grant No. DP240103070), Australian Research Council ARC Early Career Industry Fellowship (Grant No. IE230100119), Australian Research Council ARC Early Career Industry Fellowship (Grant No. IE240100275), University of Adelaide, Sustainability FAME Strategy Internal Grant 2023. 

\bibliography{aaai25}
\end{document}